\documentclass{article}
\usepackage{arxiv}
\usepackage[square,sort,comma,numbers]{natbib}
\usepackage[utf8]{inputenc} 
\usepackage[T1]{fontenc}    
\usepackage[shortlabels]{enumitem}
\usepackage[obeyspaces]{url}
\usepackage{wrapfig}
\usepackage{booktabs}       
\usepackage{amsfonts}       
\usepackage{nicefrac}       
\usepackage{microtype}      
\usepackage{graphicx}
\usepackage{natbib}
\usepackage{doi}
\usepackage{multirow}
\usepackage{tabularx}
\usepackage{tabulary}
\usepackage{makecell}
\usepackage{tabularx,ragged2e}
\usepackage{fancyhdr, blindtext}
\usepackage{subfigure}
\usepackage{float}
\usepackage{hyperref}
\usepackage{hhline}
\usepackage{array}
\usepackage{amsmath}
\usepackage{diagbox}
\usepackage{color}
\usepackage{textcomp}
\usepackage{comment}
\usepackage{relsize}
\usepackage[symbol]{footmisc}
\usepackage{caption}
\usepackage{chngcntr}
\usepackage{amssymb}%
\usepackage{amsthm}%
\usepackage{mathrsfs}%
\usepackage[title]{appendix}%
\usepackage{xcolor}%
\usepackage{textcomp}%
\usepackage{manyfoot}%
\usepackage{booktabs}%
\usepackage{algorithm}%
\usepackage{algorithmicx}%
\usepackage{algpseudocode}%
\usepackage{listings}%
\usepackage{natbib} 
\usepackage{comment}
\usepackage{etex} 
\usepackage{morewrites}
\usepackage{authblk} 
\usepackage{multicol}

\renewcommand{\headeright}{Generative AI in Surgery}
\renewcommand{\shorttitle}{SimGen: Simultaneous Image-Mask Generation}
\renewcommand{\undertitle}{Advances in Generative AI}

\setcitestyle{square}
\hypersetup{
pdftitle={SimGen},
pdfsubject={q-bio.NC, q-bio.QM},
pdfauthor={Aditya et al.},
pdfkeywords={Diffusion models, surgical image-label synthesis, paired image-mask generation, segmentation},
}

\title{SimGen: A Diffusion-Based Framework for Simultaneous Surgical Image and Segmentation Mask Generation}

\date{}

\author{ 
    \textbf{Aditya Bhat}\textsuperscript{1,†}, 
    \textbf{Rupak Bose}\textsuperscript{1,†}, 
    \textbf{Chinedu Innocent Nwoye}\textsuperscript{1,2,†,*},
    \textbf{Nicolas Padoy}\textsuperscript{1,2}\\
    \textsuperscript{1}ICube, UMR7357, CNRS, INSERM, University of Strasbourg, France\\
    \textsuperscript{2}IHU Strasbourg, France\\
    {\smaller
        \textsuperscript{†}{These authors contributed equally}.\\
        \textsuperscript{*}Corresponding author: nwoye@unistra.fr
    }
}

\begin{document}
\maketitle

\begin{abstract}
    Acquiring and annotating surgical data is often resource-intensive, ethical constraining, and requiring significant expert involvement. While generative AI models like text-to-image can alleviate data scarcity, incorporating spatial annotations, such as segmentation masks, is crucial for precision-driven surgical applications, simulation, and education.
    This study introduces both a novel task and method, \textit{SimGen}, for \textit{S}imultaneous \textit{I}mage and \textit{M}ask \textit{Gen}eration. SimGen is a diffusion model based on the DDPM framework and Residual U-Net, designed to jointly generate high-fidelity surgical images and their corresponding segmentation masks. The model leverages cross-correlation priors to capture dependencies between continuous image and discrete mask distributions. Additionally, a Canonical Fibonacci Lattice (CFL) is employed to enhance class separability and uniformity in the RGB space of the masks. 
    SimGen delivers high-fidelity images and accurate segmentation masks, outperforming baselines across six public datasets assessed on image and semantic inception distance metrics.
    Ablation study shows that the CFL improves mask quality and spatial separation. Downstream experiments suggest generated image-mask pairs are usable if regulations limit human data release for research.
    This work offers a cost-effective solution for generating paired surgical images and complex labels, advancing surgical AI development by reducing the need for expensive manual annotations.
\end{abstract}
\keywords{: Diffusion models\and  surgical image-label synthesis\and  paired image-mask generation \and generative AI \and segmentation\\[0.05in]}

\begin{figure}[ht]
  \begin{center}
    \includegraphics[width=\textwidth]{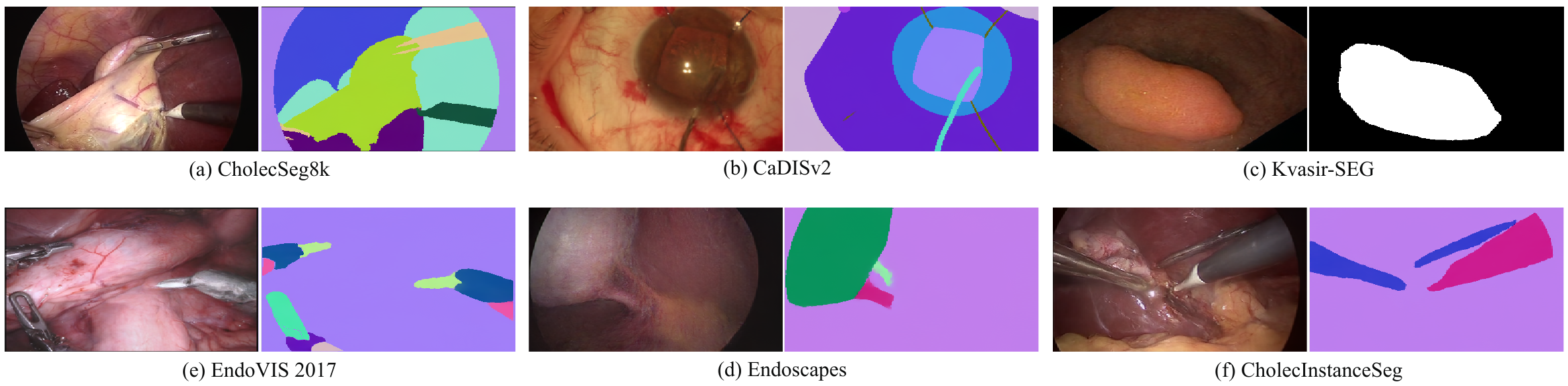}
    \caption{Sample outputs of SimGen across the 6 explored datasets. For each pair, the generated photorealistic image is on left and the generated boundary-aligned segmentation mask is on the right. The mask is colored for visualization with the Canonical Fibonacci Lattice (CFL) function.
    }
    \label{fig:qualitative_visualization}
  \end{center}
\end{figure}

\twocolumn

\section{Introduction}
Data acquisition in surgical AI presents significant challenges due to the ethical, practical, and resource-intensive demands of collecting and annotating large-scale datasets. High-quality annotations, particularly segmentation masks, require expert knowledge and precise labeling, making them both costly and time-consuming. 
Generative AI methods, such as text-to-image \cite{nwoye2024surgical,allmendinger2024navigating} and text-to-video \cite{sun2024bora,cho2024surgen}, have emerged as alternatives to address data scarcity by synthetically generating visual content.
The utility of generative AI are mostly for surgical education and simulation. In those cases, text prompts, fed as inputs to the generative methods, can serve as labels for generated images or videos, providing utility for simple tasks like illustrating surgical phases or concepts \cite{colleoni2022ssis}. However, more complex tasks—such as demonstrating anatomical boundaries, locating surgical tools, or mapping tissue structures~\cite{lou2023min,nwoye2021deep}
—demand detailed spatial annotations like segmentation masks or bounding boxes. These structured annotations cannot be adequately captured through text prompts alone, leaving a critical gap in current generative models.
Mask-to-image generation models~\cite{colleoni2024guided} partially address this gap by using segmentation masks as inputs to generate synthetic images. However, the manual creation of these masks remains prohibitively expensive and time-consuming, limiting scalability and applicability. 
This challenge is particularly evident in surgical data science, where precise spatial annotations are crucial for tasks such as surgical scene understanding~\cite{sharma2023surgical,yamlahi2023self} and tool tracking~\cite{nwoye2024surgitrack}, necessitating large-scale datasets.

To overcome the limitations of existing models, we propose \textit{SimGen} (Simultaneous Image-Mask Generator), a diffusion-based model that generates high-quality surgical images and their corresponding segmentation masks simultaneously. Built on the framework of Denoising Diffusion Probabilistic Models (DDPMs)~\cite{ho2020denoising}, 
SimGen employs cross-correlation priors on concatenated image and mask channels to capture dependencies between continuous image data and discrete mask distributions. This, combined with the Central Limit Theorem (CLT) to stabilize the noise distribution, enhances the signal-to-noise ratio, resulting in more accurate and coherent image-mask generation without adversarial training.
Additionally, SimGen employs a residual U-Net architecture that enhances gradient flow and model convergence, stabilizing the training process. This architecture, with two residual skip connections, mitigates gradient vanishing issues. It equally avoids exploding gradients by regressing to a single continuous distribution rather than a mix of continuous and discrete distributions.

A further innovation in SimGen is the integration of the \textit{Canonical Fibonacci Lattice} (CFL), an algorithm that projects segmentation class labels onto a uniformly spaced RGB unit sphere. This prevents color overlap by improving the separability of class segments in the RGB space, facilitating stable model convergence. During inference, cosine similarity between the learned masks and the Fibonacci-projected outputs is used to recover the original class labels. Furthermore, the generated masks can be easily adapted into bounding boxes using a simple coordinate extraction algorithm, offering flexibility across diverse datasets and supporting a wide range of applications.

We explore our method on six publicly available surgical datasets
\cite{hong2020cholecseg8k,allan20192017,murali2023endoscapes,jha2020kvasir,alabi2024cholecinstanceseg,luengo20212020}, spanning a variety of procedures, including laparoscopic, gastroscopic, colonoscopic, and cataract surgeries. We assess SimGen’s performance using Fréchet Inception Distance (FID) and Kernel Inception Distance (KID) for image quality. We also computed Semantic FID/KID for each semantic region based on the generated class mask to evaluate the image-mask alignment, mask class and boundary correctness. Our results show that SimGen produces higher-fidelity surgical images and more accurate segmentation masks compared to the baselines.

We evaluate SimGen's utility in downstream tasks, such as segmentation, by training a UNet model \cite{zhang2018road} on generated image-mask pairs. Our results demonstrate that SimGen produces meaningful images alongside boundary-aligned masks that deep learning models could learn from. Although UNet trained on synthetic data underperforms compared to real data (52.1\% vs. 41.7\%), which is expected, this moderate performance underscores SimGen's value when real human data cannot be released due to ethical or regulatory concerns. Further analysis, where SimGen is trained on a superset of available real data, shows improvements, demonstrating that smaller institutions with limited datasets can greatly benefit from SimGen trained on data from larger institutions.
The observed distribution shift between real and synthetic data also highlights SimGen's potential for transfer learning and domain adaptation studies. 
We also demonstrate that SimGen's generated image-mask pairs exhibit superior alignment compared to using state-of-the-art SegFormer \cite{xie2021segformer} to predict masks from generated images in surgical simulation.

In summary, our contributions are as follows:
\begin{enumerate}
    \item We introduce a novel task of \textit{paired image-mask generation}, addressing the need for both visual data and structured annotations.
    \item We propose \textit{SimGen}, a diffusion-based model that implicitly utilizes cross-correlation priors to jointly generate high-quality surgical images and segmentation masks without the need for adversarial training.
    \item We integrate \textit{Canonical Fibonacci Lattice} to enhance segmentation mask separability in RGB space, preventing class overlap and improving training stability.
    \item We introduce Semantic Inception Distance (SID), a metric for evaluating fidelity, paired alignment, boundary accuracy, and class correctness in generated image-mask pairs.
    \item We demonstrate SimGen's superior performance through extensive evaluation on five public surgical datasets, achieving state-of-the-art results in both image and image-mask alignment fidelity.
\end{enumerate}

\section{Related Works}

\subsection{Image Generation} 
Early image generation models like Variational Autoencoders (VAEs)~\cite{kingma2013auto} introduced latent space learning, but struggled with producing high-quality images. The introduction of Generative Adversarial Networks (GANs)~\cite{goodfellow2014generative} marked a significant breakthrough, with advances such as conditional~\cite{mirza2014conditional} and Style GANs~\cite{karras2019style}, etc., 
enabling more controlled image generation. However, GANs face issues like mode collapse and instability, limiting their robustness across tasks.

\subsection{Diffusion Models} Diffusion models, such as DDPM~\cite{ho2020denoising}, resolve issues faced by GANs using a likelihood-based loss function, resulting in a more stable model capable of generating diverse, high-quality images. Denoising Diffusion Implicit Models (DDIMs)~\cite{song2020denoising}, improved sampling efficiency of DDPM. Latent~\cite{rombach2022high},
conditional~\cite{ho2022classifier},
and stable~\cite{rombach2022high} diffusion models power modern text-to-image (e.g. Imagen~\cite{saharia2022photorealistic}) and text-to-video (e.g. Sora~\cite{videoworldsimulators2024}) generation, offering more versatility in content creation.

\subsection{Surgical Data Synthesis}
Given the challenges of data acquisition in the surgical domain, generative models such as Surgical Imagen~\cite{nwoye2024surgical}, and counterparts~\cite{allmendinger2024navigating} explored text-to-image generation, whereas Bora~\cite{sun2024bora}, SurGen~\cite{cho2024surgen} and Endora~\cite{li2024endora} extend this to videos. These models utilize text prompts describing surgical phases and tool-tissue interactions~\cite{nwoye2022rendezvous} to create realistic surgical contents. 
Aside the limited precision in the input text prompt, its utility alongside generated outputs in complex applications tasks such as simulation~\cite{colleoni2022ssis} fall short in providing the necessary overlays, underscoring the need for generating paired image-mask data, a gap this work seeks to fill.


\section{Methods}
Our objective is to generate paired images and segmentation masks directly from noise, ensuring they are optimized for high-impact surgical applications.

\subsection{Task Formalization}
Given a dataset $D=\{(x_i,y_i)\}_{i=1}^n$ such that $y_i = f(x_i)$ with $x_i \in \mathbb{R}^{3 \times h \times w},  y_i \in \mathbb{R}^{c \times h \times w}$ where $c$ is the number of channels in the sample $y_i$ and the function $f: x_i \rightarrow y_i$ maps the input $x_i$ to $y_i$. 
Our aim is to develop a model $\mathcal{M}$ that generates ($\bar{x}, \bar{y}$) such that $\bar{y_i} = f(\bar{x_i})$ and the likelihood for $\bar{x}$ and $\bar{y}$ to be similar to samples drawn from $D$ is maximized.
In our case, ($\bar{x}, \bar{y}$) represents paired image and segmentation mask.

\begin{figure}[t]
\centering
\includegraphics[width=.7\linewidth]{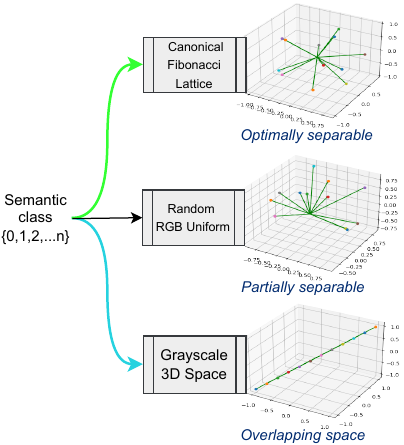}
\caption{Illustration of the proposed Fibonacci projection of semantic class identities to 3D hyperspace in comparison with random RGB and grayscale spaces.}
\label{fig:cfl}
\end{figure}

\subsection{Canonical Fibonacci Lattice (CFL)}
First, we implement CFL (Fig.~\ref{fig:cfl})  to uniformly distributes points on a unit sphere, maximizing class separation in high-dimensional spaces. This addresses the challenge of class overlap by projecting classes onto a unit 3D sphere using golden angles and the Fibonacci sequence. We observed that CFL facilitates stable model convergence, without which the generated mask lacks class separability.
The pseudocode guiding the implementation of the CFL for class separability in the mask generation in presented in Algorithm \ref{alg:cfl}.

\begin{algorithm}
\caption{Canonical Fibonacci Lattice for Point \(i\)}
\label{alg:cfl}
\begin{algorithmic}[1]
\State \textbf{Input:}  \( NC, i \) \# for number of classes
\State \textbf{Output:} 3D coordinates \( (x_i, y_i, z_i) \)

\State Set golden ratio \( GR \gets \frac{1 + \sqrt{5}}{2} \)
\State Set golden angle \( \theta \gets \frac{2 \pi i}{GR} \)
\State Set \( \phi \gets \arccos\left( 1 - 2 \cdot \frac{i + 0.5}{NC} \right) \)

\State Compute \( x \gets \cos(\theta) \cdot \sin(\phi) \)
\State Compute \( y \gets \sin(\theta) \cdot \sin(\phi) \)
\State Compute \( z \gets \cos(\phi) \)

\State \textbf{Return} \( (x, y, z) \)
\end{algorithmic}
\end{algorithm}

\begin{figure*}[t]
\centering
\includegraphics[width=.9\textwidth]{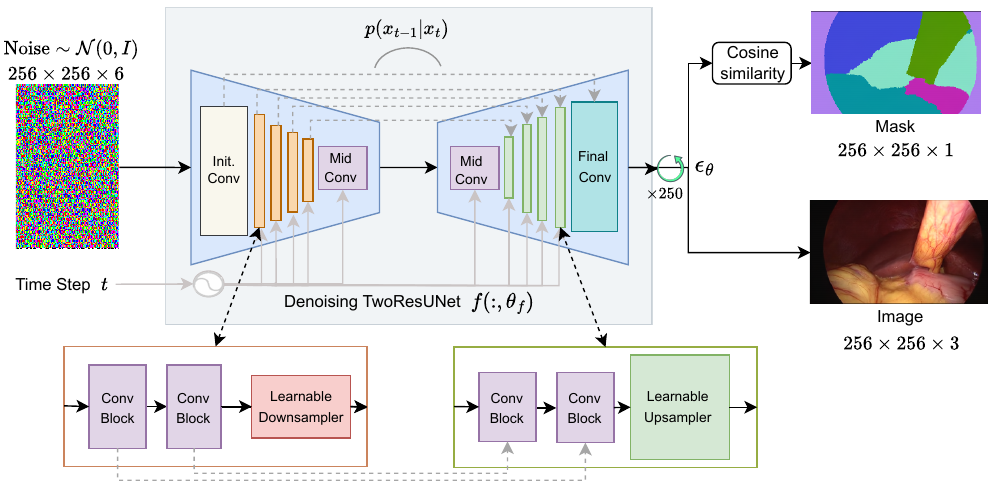}
\caption{Architecture of the proposed SimGen model showing its several key components that work together to generate paired image-mask from noise.}
\label{fig:model}
\end{figure*}

\subsection{Model Logic and Architecture}
Our model is based on a ResUNet architecture and operates within the framework of a diffusion model (DDPM) \cite{ho2020denoising}. This denoising ResUNet model $f(:,\theta_f)$ employs a symmetrical encoder-decoder pipeline, specifically designed to process noisy image-mask paired input $(x_t, y_t)$, with noise level $t$, and accurately reconstruct the corresponding output pair following the DDPM reverse process.

During the forward diffusion process, Gaussian noise $\mathcal{N}(0,1)$ is incrementally added to the concatenated 6-channel input ($x_0$ $\oplus$ $y_0$) over a series of timesteps. By the terminal timestep, this noise approximates a Gaussian distribution due to the Central Limit Theorem (CLT), which asserts that the sum of independent random variables converges to a normal distribution, regardless of initial distributions. SimGen capitalizes on this property to stabilize the noise distribution, enhancing the signal-to-noise ratio and enabling the model to focus on learning structured information rather than struggling with irregular noise patterns. This design avoids the complexities of adversarial training, ensuring smoother convergence and higher-quality image-mask generation.

To further refine the noise removal process, cross-correlation priors are applied to the concatenated input ($x_t \oplus y_t$), explicitly modeling the dependencies between continuous image data and discrete mask distributions. This cohesive noise handling improves the model's ability to process and generate consistent image-mask pairs. Additionally, the Two-residual U-Net architecture incorporates stabilized gradient flow through skip connections, enabling the network to regress effectively to the learned noise distribution. 
These connections mitigate gradient vanishing issues by stabilizing gradient flow and ensuring efficient feature propagation across the network.

The architecture, as illustrated in Fig. \ref{fig:model}, starts with an \textit{initial convolution} that projects the noisy 6-channel input ($x_t \oplus y_t$) into a feature space while preserving spatial resolution for decoder use through residual connections. 
A \textit{time step embedding} module encodes the temporal step ($t$) into a higher-dimensional space and combines it with the feature map to capture both spatial and temporal dynamics. Each layer contains two \textit{ConvBlock} modules, where each ConvBlock processes features using GroupNorm, Conv2D, and GELU. Residual connections are made from each ConvBlock in every layer, linking the encoder to the decoder for high-resolution recovery. A \textit{learnable downsampler} reduces spatial resolution while expanding spectral resolution before passing features through the bottleneck (\textit{Mid Conv.}). The decoder uses a \textit{learnable upsampler} to restore spatial resolution with transposed convolutions, utilizing the residual connections to produce a final 6-channel noise prediction ($\epsilon_\theta$).

\subsection{Training and Inference}

We utilize the DDPM for sampling, with the diffusion loss defined as the mean squared error (MSE) between the predicted noise \( \epsilon_\theta(X_t, t) \) and the true noise \( \epsilon \) for a noisy image \( X_t \) at time step \( t \): 
$
\mathcal{L}_{\text{diffusion}} = \mathbb{E}_{t, \epsilon}\left[\|\epsilon - \epsilon_\theta(X_t, t)\|^2\right].
$
During inference, we compute cosine similarity to align generated pixel values with the color vectors in the CFL dictionary, ensuring accurate class assignment based on the highest similarity.

\begin{table*}[!htpb]
    \centering
    \caption{Dataset used for training image-mask pair generation models.}
    \label{tab:dataset}    
    \setlength{\tabcolsep}{12pt}    
    \resizebox{0.999\linewidth}{!}{%
    \begin{tabular}{@{}lccccc@{}}
    \toprule
        Dataset & Procedure & \# Images & Mask Type & \# Classes & Train / Val /Test \\
        \midrule
        CholecSeg8K \cite{hong2020cholecseg8k} & Laparoscopy& 8080 & semantic & 13 & 5600~/~X~/~2480\\
        CaDISv2 \cite{grammatikopoulou2021cadis} & Cataract Surgery& 4670 & semantic & 36 & 3550~/~534~/~586\\
         EndoVIS \cite{allan20192017}&  Robotic Laparoscopy& 1800 & instruments & 8 & 1800~/~X~/~750\\
         EndoVIS \cite{allan20192017}&  Robotic Laparoscopy & 1800 & parts & 4 & 1800~/~X~/~750\\
        Kvasir-SEG\cite{jha2020kvasir} & Colonoscopy& 1000 & binary & 2 & 880~/~120~/~X\\
        Endoscapes\cite{murali2023endoscapes}& Laparoscopy& 493 & semantic & 7 & 343~/~76~/~74\\
        CholecInstanceSeg\cite{alabi2024cholecinstanceseg} & Laparoscopy& 41933 & semantic & 8 & 26830~/~3804~/~11299 \\
        \bottomrule
    \end{tabular}
    }
\end{table*}

\subsection{Evaluation on Semantic Inception Distance (SID)}
To evaluate the quality of the generated image-mask pairs, we introduce \textit{Semantic Inception Distance (SID)}, an extension of the traditional Inception Distance (ID) that evaluates generated image correctness by focusing on specific semantic regions within an image. SID builds on existing ID metrics like Fréchet Inception Distance (FID) and Kernel Inception Distance (KID), using them to compute the distance for individual semantic regions, such as anatomical structures and tools, by leveraging segmentation masks. We refer to these extensions as semantic FID (sFID) and semantic KID (sKID).

While traditional ID metrics measure global distribution similarity, SID isolates semantic regions to assess boundary alignment, class assignment correctness, and anatomical fidelity more effectively. 
The key concepts in SID includes:

\begin{enumerate} 

\item \textit{Region Isolation}: Unlike traditional FID/KID, SID uses segmentation masks to crop and compare each semantic region within the image, enabling finer-grained comparison and more precise assessments.

\item \textit{Boundary Alignment:} 
 Focusing on individual semantic regions, SID directly penalizes any misalignment in the boundaries of object structures within an image. Misaligned regions (e.g., incorrect cropping or overlap of adjacent organs or tools) will result in a worse similarity score for that region, emphasizing the importance of accurate boundary alignment, which traditional ID metrics do not capture.

\item \textit{Class Assignment Correctness:}
SID ensures that each region is correctly classified. For instance, if a mask for the liver overlaps with a region designated for the gallbladder, SID will yield a significantly worse similarity score for that region, thus capturing class assignment errors that would otherwise be overlooked by global metrics. This assesses the image-mask alignment correctness.
 
\item \textit{Exhaustive Element Evaluation:}
SID supports assessment of both anatomical structures and tools and any object of interest in the generated surgical image. In our case, objects of the same class are evaluated collectively - this can be extended to instance-based assessment. For text-conditioned generation, SID could further align element classes with input text keywords for enhanced semantic coherence.
 
\end{enumerate}


\section{Experiments}

\subsection{Dataset}
Six surgical segmentation datasets were explored: 
CaDISv2~\cite{luengo20212020}, Cholecseg8k~\cite{hong2020cholecseg8k},  CholecInstanceSeg~\cite{alabi2024cholecinstanceseg}, Endoscapes~\cite{murali2023endoscapes},  EndoVIS 2017~\cite{allan20192017}, and Kvasir-SEG~\cite{jha2020kvasir}.
We train our proposed and baseline models on the training split of each dataset. Downstream task evaluations are performed on the test splits. More dataset details are provided in Table \ref{tab:dataset}.

\subsection{Implementation Details}
We trained SimGen on inputs of size $256 \times 256$ with 64 feature maps, 250 diffusion timesteps, and (1, 2, 4, 8) multipliers for features. The 62.7M parameter SimGen model is trained on a V100 32GB GPU for 5 days per dataset using a batch size of 16, with the Adam optimizer having a learning rate of \(1.0 \times 10^{-4}\). SimGen's implementation is based on PyTorch.

\subsection{Baselines}
We selected 3 generative models that are close our tasks as baselines:
DCGAN~\cite{radford2015unsupervised}, Pix2Pix~\cite{isola2017image}, and Convolutional VAE~\cite{kingma2013auto}. 
Each was trained to full saturation, with no further performance improvements observed.

\subsection{Evaluation Protocols}
We assess the performance of our method using several metrics: (1) image FID score to assess the fidelity of generated images compared to real data; given the small size of training data, we also compute the KID score, (2) SID (sFID and sKID) to evaluate the image-mask boundary alignment and tool/anatomical class correctness, (3) Intersection over Union (IoU) from downstream segmentation model (ResUNet \cite{zhang2018road}) trained on varying percentage compositions of real and generated data, and (4) comparisons with baselines.


\section{Results}

\subsection{Generated Image Quality}
We evaluate the generated image fidelity in comparison with the real image distribution. As shown in Table \ref{tab:fid}, 
using FID metric, which estimates at a global image feature scale, SimGen obtains a significantly lower FID compared to the baseline models, though it still shows room for improvement when compared to the real data. At a local feature scale, the KID score of SimGen approximates the real images and fairs better than all the baselines across the compared datasets.
Note that we show baseline performances only for complementary datasets without re-training them for datasets like Endoscape, CholecInstanceSeg, etc., which belongs to the same source as CholecSeg8k.

The qualitative results highlights the true fidelity and photorealistic nature of SimGen's generated image-mask pairs for six dataset as presented in Fig. \ref{fig:qualitative_visualization}.

\begin{table*}[!thbp]
    \centering   
    \caption{Automatic evaluation of generated images using mean FID and KID scores across six datasets:Cataract Segmentation (CaDIS), CholecSeg8k (CSeg8K), CholecInstanceSeg (CISeg), Endoscapes (EdScp), EndoVIS, and Kvasir.}
    \label{tab:fid} 
    \setlength{\tabcolsep}{4pt}    
    \resizebox{0.998\linewidth}{!}{%
    \begin{tabular}{lccccccrccccccc}
        \toprule
        \multirow{2}{*}{Models} & \multicolumn{6}{c}{FID~$\downarrow$} & \phantom{abc} & \multicolumn{6}{c}{KID~$\downarrow$}\\
        \cmidrule{2-7}\cmidrule{9-14}
        & CaDIS & CSeg8k & CISeg & EdScp & EndoVIS & Kvasir && CaDIS & CSeg8k & CISeg & EdScp & EndoVIS & Kvasir &\\
        \midrule
        Pix2Pix~\cite{isola2017image} & 414.1 & 432.9 & - & - & - & 425.9 && 0.44 & 0.54 & - & - & - & 0.48 & \\
        DCGAN~\cite{radford2015unsupervised} & 351.1 & 466.7 & - & - & - & 417.3 && 0.44 & 0.63 & - & - & - & 0.46 & \\
        VAE & 432.1 & 403.1 & - & - & - & 341.7 && 0.64 & 0.55 & - & - & - & 0.50 & \\
        SimGen (ours) & 74.2 & 57.5 & 53.8 & 217.9 & 107.8 & 136.9 && 0.08 & 0.04 & 0.05 & 0.23 & 0.10 & 0.16 & \\
        \midrule
        Real data & 13.3 & 5.0 & 2.1 & 72.7 & 29.8 & 43.1 && 0.00 & 0.00 & 0.00 & 0.00 & 0.00 & 0.00 & \\
        \bottomrule
    \end{tabular}
    }
\end{table*}

\begin{figure*}[!htbp]
    \centering
    \includegraphics[width=0.99\linewidth]{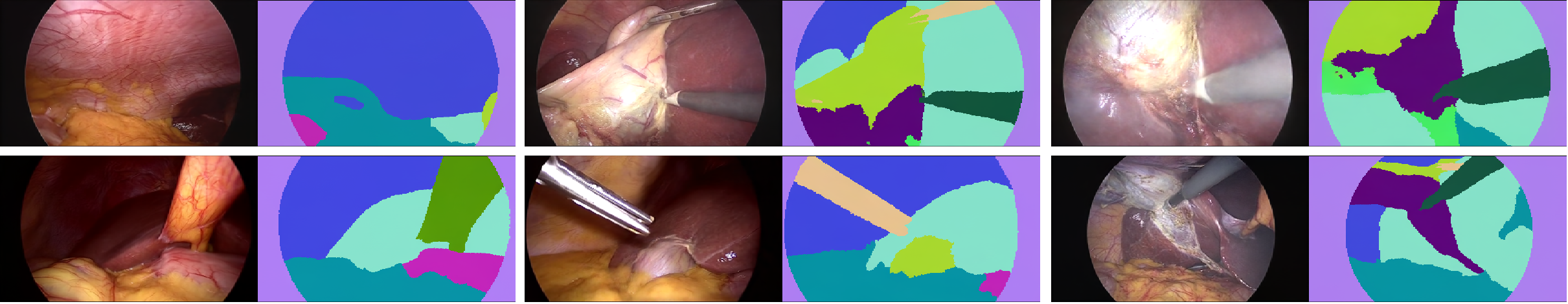}
    \caption{A random set of 6 generated image-mask pairs from the CholecSeg8K dataset}
    \label{fig:cholecseg8k-appendix}
\end{figure*}

\begin{figure*}[!htbp]
    \centering
    \includegraphics[width=0.99\linewidth]{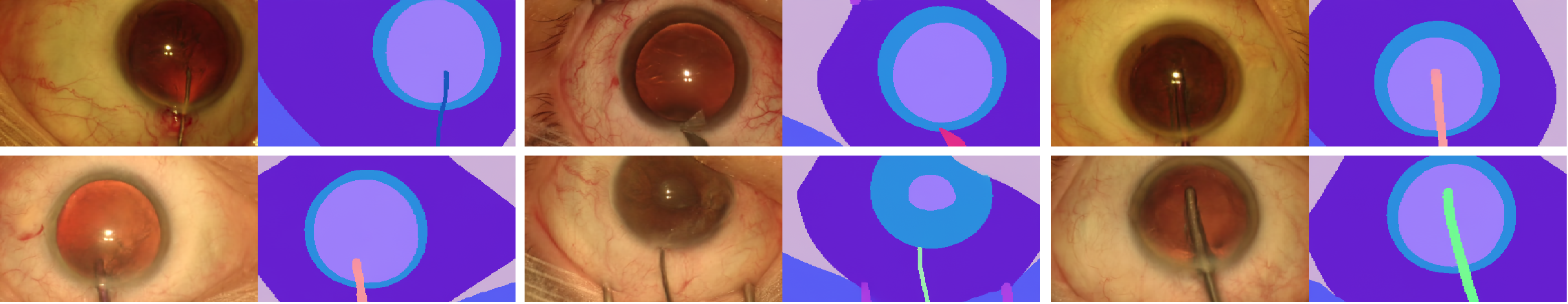}
    \caption{A random set of 6 generated image-mask pairs from the CaDISv2 dataset}
    \label{fig:cadisv2-appendix}
\end{figure*}

\begin{figure*}[!htbp]
    \centering
    \includegraphics[width=0.99\linewidth]{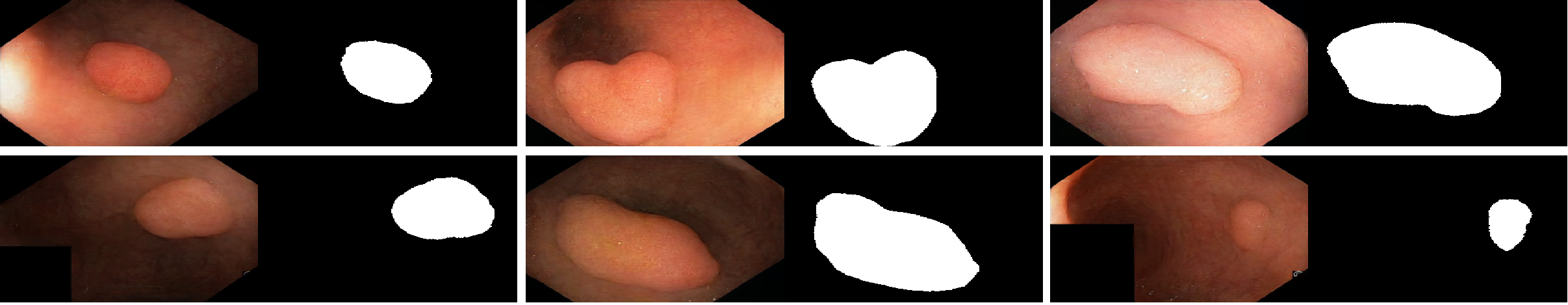}
    \caption{A random set of 6 generated image-mask pairs from the Kvasir-SEG dataset}
    \label{fig:kvasir-appendix}
\end{figure*}

\begin{figure*}[!htbp]
    \centering
    \includegraphics[width=0.99\linewidth]{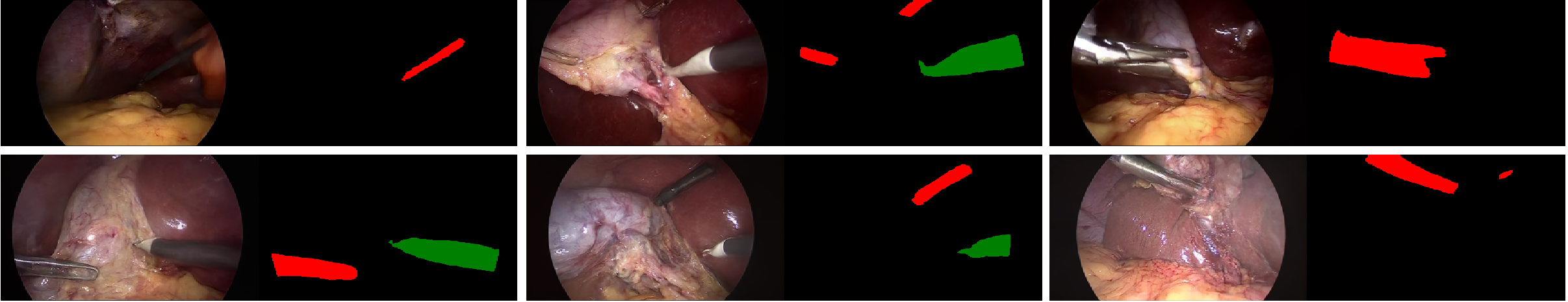}
    \caption{A random set of 6 generated image-mask pairs from the CholecInstanceSeg dataset}
    \label{fig:cholecinstanceseg-appendix}
\end{figure*}

\begin{figure*}[!htbp]
    \centering
    \includegraphics[width=0.99\linewidth]{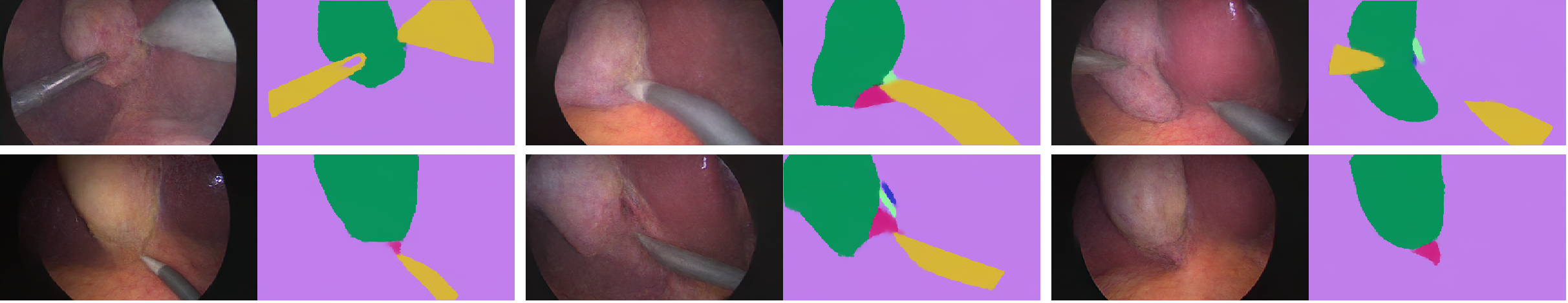}
    \caption{A random set of 6 generated image-mask pairs from the Endoscapes dataset}
    \label{fig:endoscapes-appendix}
\end{figure*}

\begin{figure*}[!htbp]
    \centering
    \includegraphics[width=0.99\linewidth]{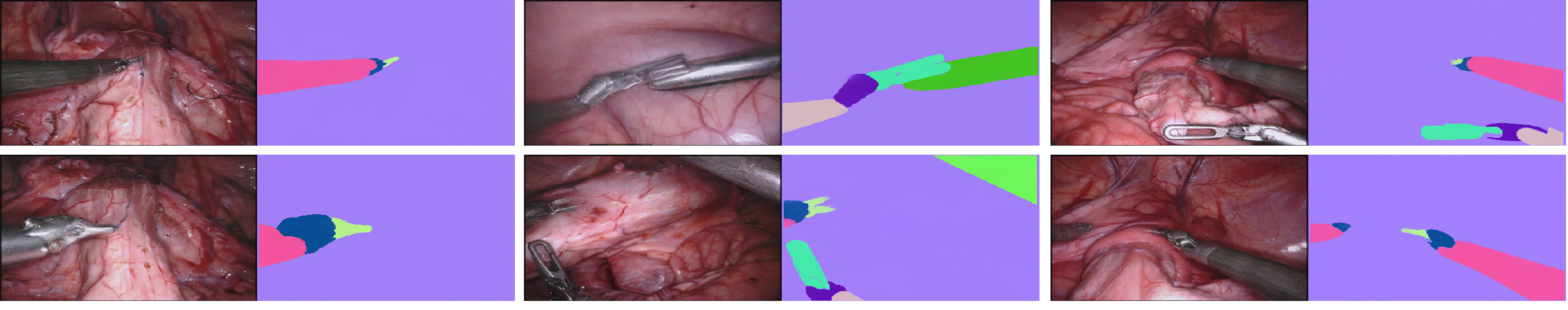}
    \caption{A random set of 6 generated image-mask pairs from the EndoVIS dataset}
    \label{fig:endovis-appendix}
\end{figure*}

\begin{table*}[!th]
    \centering
    \caption{Automatic evaluation of generated image-mask alignment using mean sFID and sKID scores across six datasets: Cataract Segmentation (CaDIS), CholecSeg8k (CSeg8K), CholecInstanceSeg (CISeg), Endoscapes (EdScp), EndoVIS, and Kvasir.}
    \label{tab:cfid} 
    \setlength{\tabcolsep}{4pt}    
    \resizebox{0.998\linewidth}{!}{%
    \begin{tabular}{lccccccrccccccc}
        \toprule
        \multirow{2}{*}{Models} & \multicolumn{6}{c}{Mean Per-Class sFID~$\downarrow$} & \phantom{abc} & \multicolumn{6}{c}{Mean Per-Class sKID~$\downarrow$}\\
        \cmidrule{2-7}\cmidrule{9-14}
        & CaDIS & CSeg8k & CISeg & EdScp & EndoVIS & Kvasir && CaDIS & CSeg8k & CISeg & EdScp & EndoVIS & Kvasir &\\
        \midrule
        Pix2Pix~\cite{isola2017image} & 298.1 & 329.0
        & - & - & - & 329.7 && 0.34 & 0.38 & - & - & - & 0.33 & \\
        DCGAN~\cite{radford2015unsupervised} & 257.0 & 276.0 & - & - & - & 362.6 && 0.28 & 0.29 & - & - & - & 0.35 & \\
        VAE & 277.2 & 282.5 & - & - & - & 351.1 && 0.35 & 0.34 & - & - & - & 0.48 & \\
        SimGen (ours) & 171.4 & 133.4 & 116.4 & 188.5 & 185.6 & 101.5 && 0.17 & 0.14 & 0.08 & 0.16 & 0.17 & 0.09 & \\
        \midrule
        Real data & 46.7 & 13.0 & 4.92 & 82.4 & 34.09 & 40.0 && 0.00 & 0.00 & 0.00 & 0.00 & 0.00 & 0.00 & \\
        \bottomrule
    \end{tabular}
    }
\end{table*}

\subsection{Image-Label Alignment}
Fig. \ref{fig:qualitative_visualization} also demonstrate precise alignment of the generated masks with their images. 
Using the SID metrics, the mean semantic FID and semantic KID scores showcase the agreement of the generated images in comparison with the real data as shown in Table \ref{tab:cfid}.
Our proposed model obtained better FID and KID scores compared to the baselines across the evaluated datasets.

\begin{figure*}[t]
    \centering
    \includegraphics[width=0.88\linewidth]{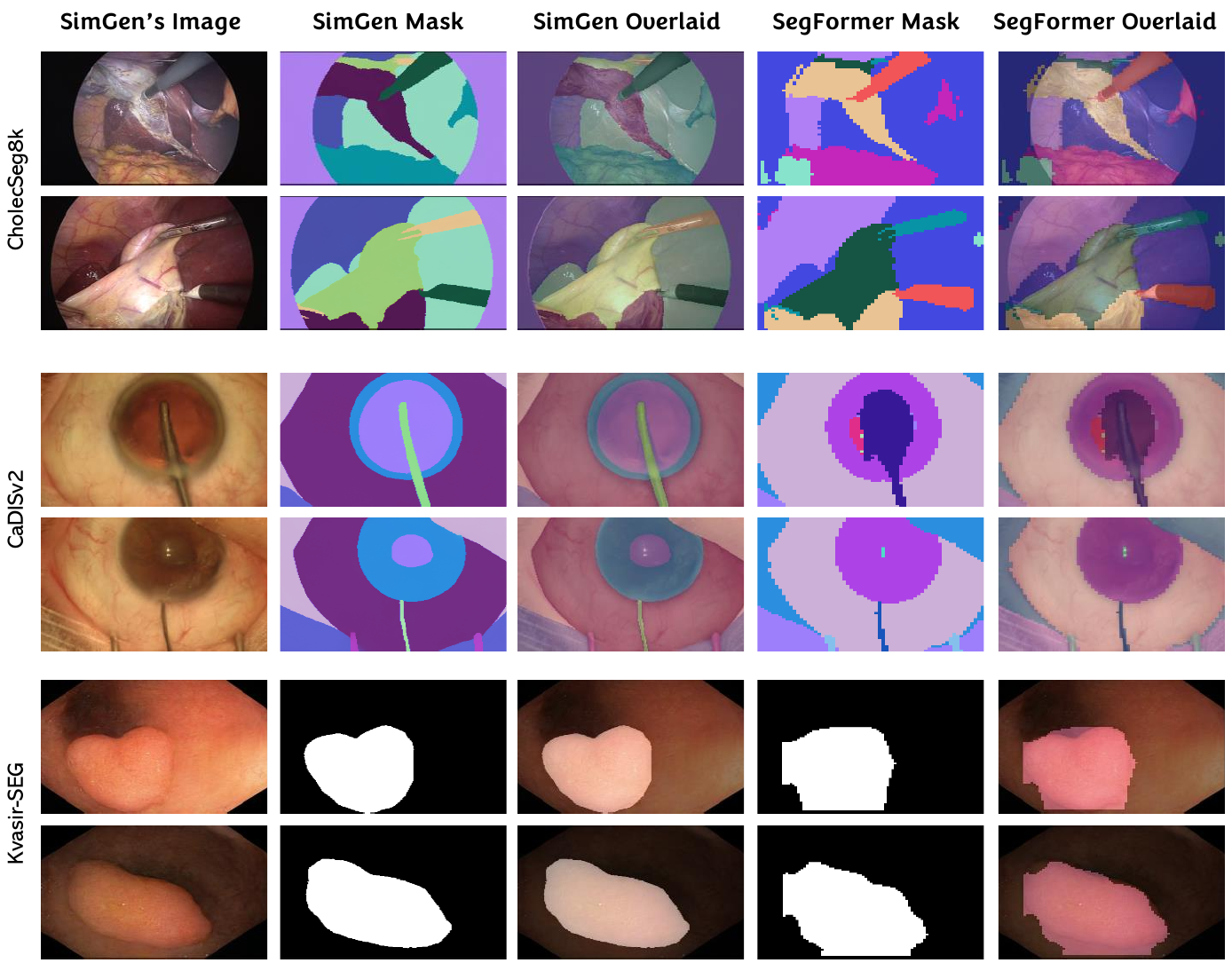}
    \caption{Comparison of anatomical semantic and boundary correctness: SimGen-generated masks versus SegFormer-predicted masks overlaid on the generated images.}
    \label{fig:overlay}
\end{figure*}

Additionally, we demonstrate that SimGen achieves superior image-mask alignment compared to using third-party segmentation models in surgical simulations. As shown in Fig. \ref{fig:overlay}, SimGen's generated masks outperform those produced by SegFormer \cite{xie2021segformer}, a state-of-the-art segmentation model which we trained on three complementary datasets explored in this work, offering better quality and alignment.

\begin{table}[t]
    \centering
    \caption{Ablation study on the impact of Canonical Fibonnaci Lattice (CFL).}
    \label{tab:ablation}
    \setlength{\tabcolsep}{13pt}    
    \resizebox{0.95\columnwidth}{!}{%
    \begin{tabular}{@{}lcc@{}}
        \toprule
        Ablation & \phantom{abc} & sFID $\downarrow$ \\
        \midrule
         RGB Mask without CFL && 183.9\\
         Final Model (RGB Mask + CFL) && 133.4\\
         \bottomrule
    \end{tabular}
    }
\end{table}

\subsection{Ablation Study}
\begin{figure}[t]
    \centering
    \includegraphics[width=0.998\linewidth]{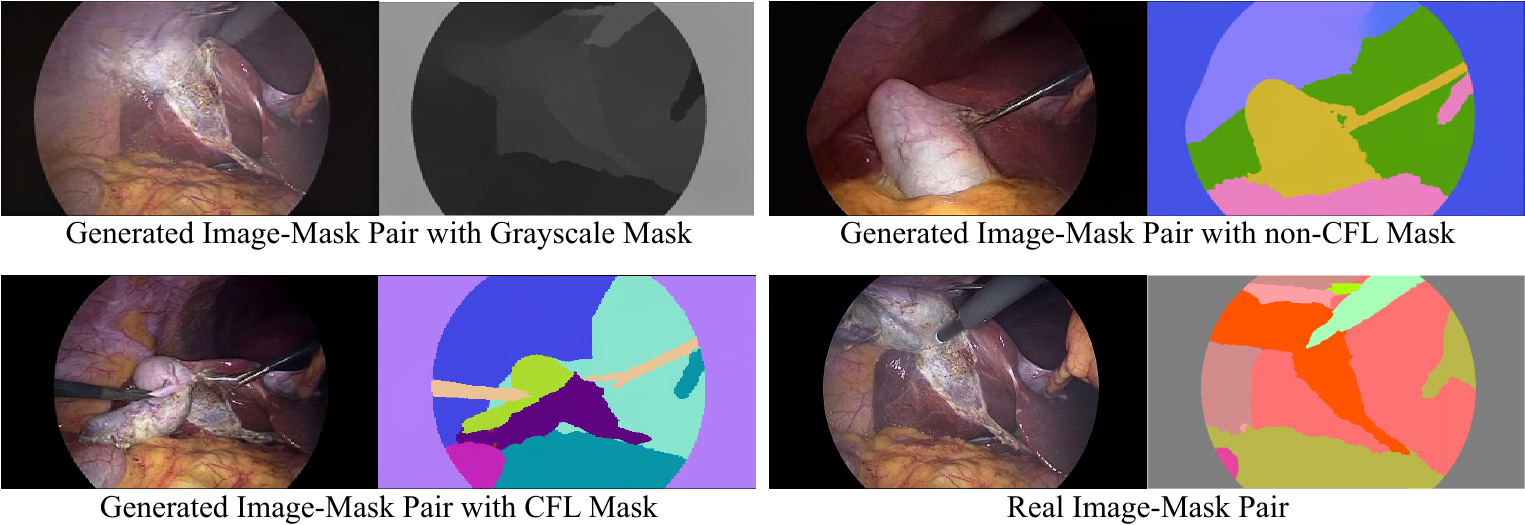}
    \caption{Qualitative ablation on the impact of CFL using CholecSeg8k dataset.}
    \label{fig:featablationcfl}
\end{figure}

To justify our use of 3D unit sphere harnessed by CFL, we conduct ablation studies on the alternative approaches in in Table \ref{tab:ablation}. 
Our proposed CFL yields the best mean class-based FID score justifying the importance of projecting mask to continuous higher dimension and the utility of uniformly spaced points on a 3D unit hyperspace. 
With increasing number of classes, CFL mitigates inherent performance drops by enhancing class separability and convergence.
As can also be seen in Fig. \ref{fig:featablationcfl}, computing class FID for grayscale masks was ineffective due to the proximity of pixel values, causing the cosine similarity function to fail in class separation and often mapping all pixels to a single class.

\begin{figure}[t]
    \centering
    \includegraphics[width=0.998\linewidth]{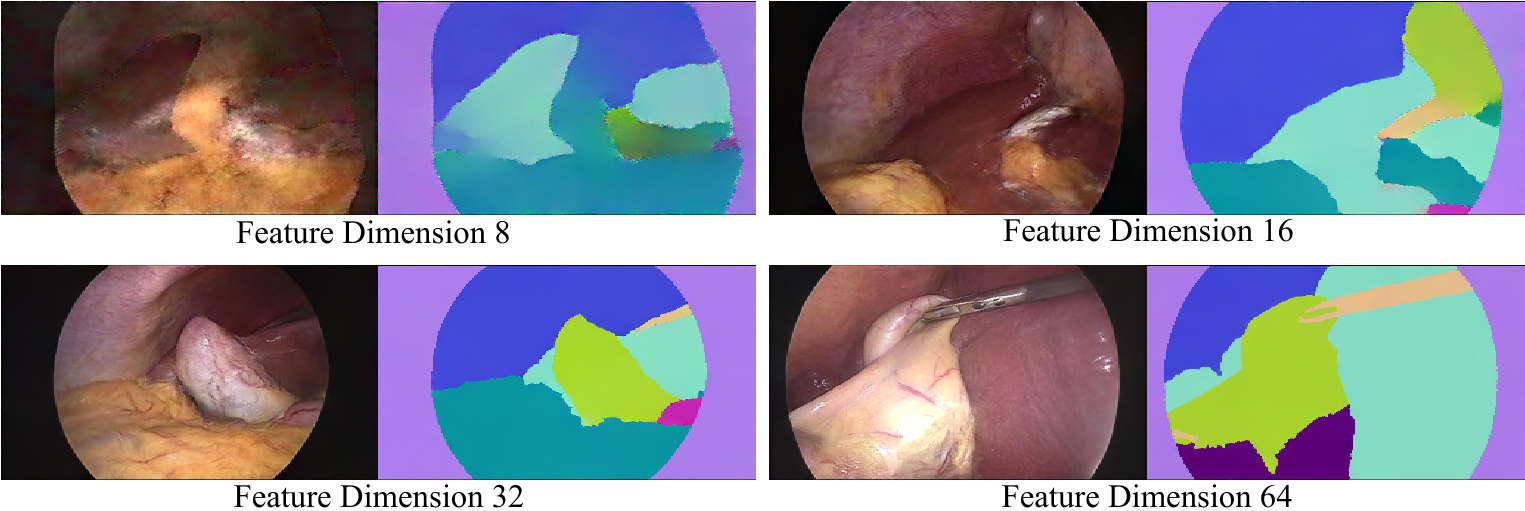}
    \caption{Qualitative ablation on feature dimension reduction using CholecSeg8k dataset.}
    \label{fig:featablation}
\end{figure}

We qualitatively ablate the effect of feature size governing the bottleneck of our model in Fig. \ref{fig:featablation}. 
From the obtained results on CholecSeg8k, we observe that increasing the feature size from 8 to 64 yields higher fidelity image-mask pairs at a given number of iterations.

\begin{figure*}[t]
    \centering
    \includegraphics[width=0.99\linewidth]{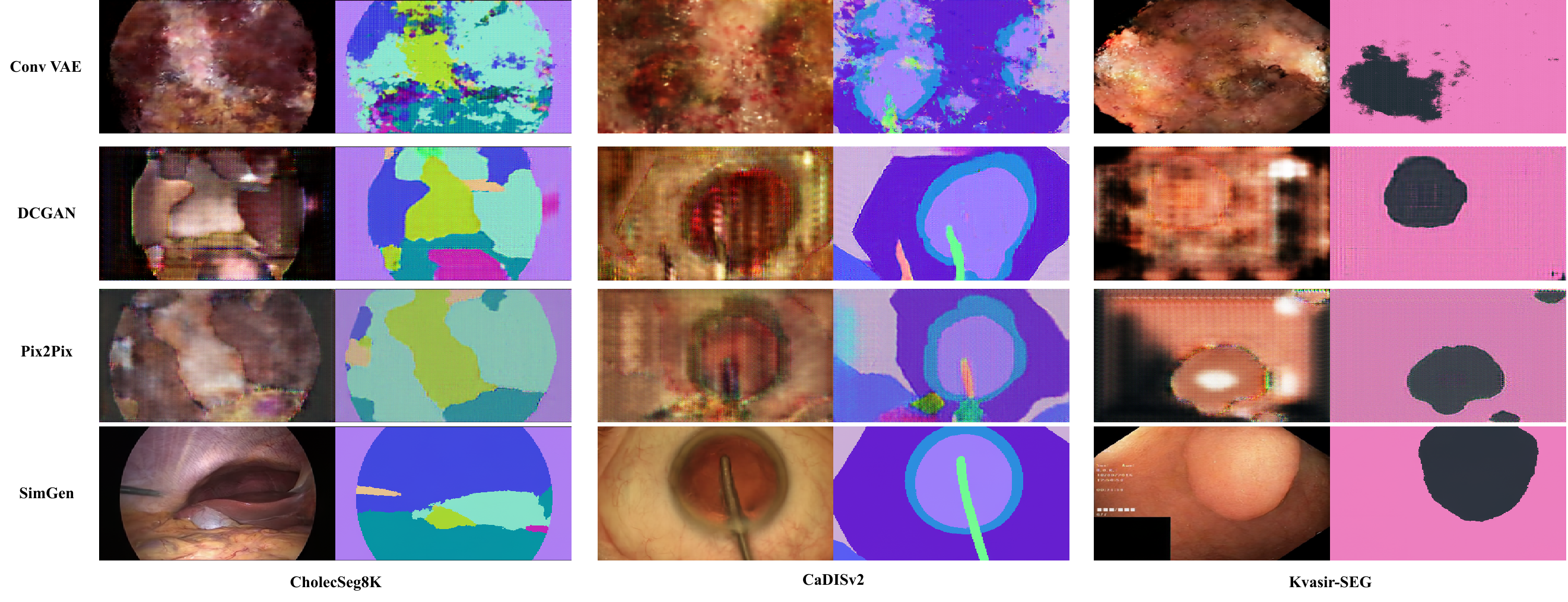}
    \caption{Qualitative results of SimGen across 3 complementary datasets in comparison with baselines.}
    \label{fig:comparison}
\end{figure*}

\subsection{Comparison with Existing Models}
Our proposed SimGen model outperforms the baselines in both Image and Class FID/KID metrics, as shown in Tables \ref{tab:fid} and \ref{tab:cfid}.
Furthermore, Fig. \ref{fig:comparison} highlights qualitative comparisons, where VAE-based models fail to generate coherent image-mask pairs likely due to mismatched continuous-discrete distributions.
DCGAN \cite{radford2015unsupervised} in its greedy approach converges on the mask generation but struggles with diverging image quality.
Pix2Pix \cite{isola2017image} attempts uniform denoising for both image and mask generation, yet with precision-lacking outputs. The baselines were trained to their performance saturation.
In contrast, SimGen excels in generating photorealistic images with well-defined, boundary-aligned masks, highlighting its superior ability to handle complex image-mask generation tasks.

\subsection{Downstream Utility}  
\begin{figure}[t]
    \centering
    \includegraphics[width=0.5\linewidth]{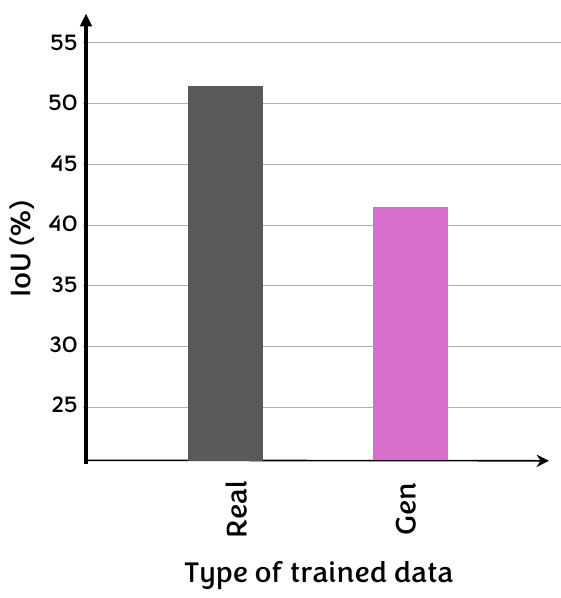}
    \caption{SimGen's utility on Segmentation. Model trained on SimGen's generated data versus real data}
    \label{fig:downstream1}
\end{figure}

Although generative models are primarily used for human training for their perceptual-focused outputs, evaluating SimGen's downstream utility in training AI is interesting. Here, we train UNet~\cite{zhang2018road} on real versus generated image-mask data for segmentation. As shown in Fig. \ref{fig:downstream1}, training solely on generated data achieves approximately 42\% IoU on the real test set. While UNet trained on entirely real data (52.1\%) outperforms training on fully synthetic data, SimGen proves valuable when real data cannot be used due to regulatory or ethical constraints.

\begin{figure*}[h]
    \centering
    \includegraphics[width=0.8\linewidth]{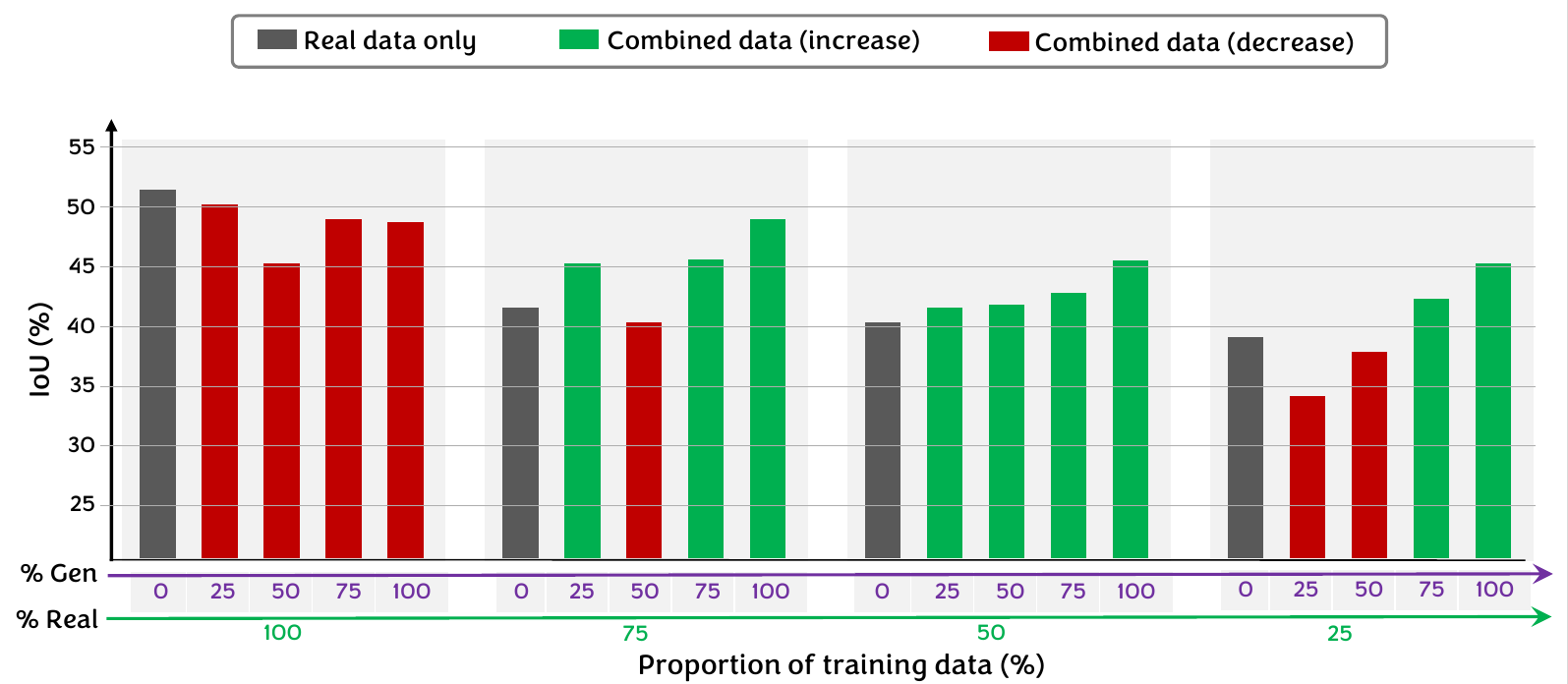}
    \caption{Utility of SimGen generated data when trained on superset dataset with more data samples.}
    \label{fig:downstream2}
\end{figure*}

The complementary use of generated data with real data was also analyzed in Fig. \ref{fig:downstream2}. If the real dataset is fully available for training, adding generated data does not improve performance and can even hinder it. This is expected, as duplicating real data in deep learning does not introduce new learning opportunities and if such duplicate is a generated sample (less quality than real data), it can introduce noise, reducing model performance. However, when the generative model is trained on a larger superset of real data, training on a smaller subset of real data (e.g., 25\%, 50\%, or 75\%) complemented with superset generated data significantly improves performance, as shown in Fig. \ref{fig:downstream2}. This improvement is due to the generative model introducing novel features not available in the real training set.

\begin{figure}[h]
    \centering
    \includegraphics[width=0.9\linewidth]{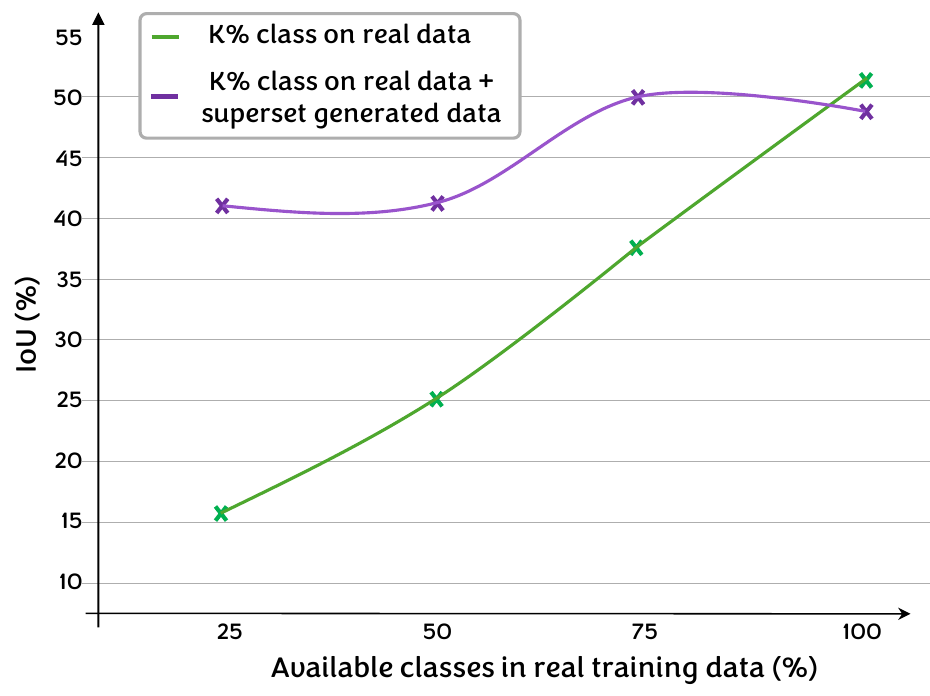}
    \caption{Utility of SimGen generated data when trained on superset dataset with more semantic class labels.}
    \label{fig:downstream3}
\end{figure}

Furthermore, In scenarios where the available real data has fewer class labels than the generative model's training data, incorporating generated data from a broader class set can substantially boost performance. As shown in Fig. \ref{fig:downstream3}, when real data is complemented with generated samples covering a broader range of classes, model performance improves by up to 20\%. 
This highlights the potential of generative models to level the playing field, enabling resource-constrained institutions to achieve better training results by leveraging synthetic data from SimGen trained on more extensive and diverse real-world datasets.

Moreso, the worsening performance in the use of generated data when the real data samples or class size matches the generated ones could be attributed to the observed distribution shift between real and synthetic data. This highlights SimGen's potential for domain adaptation research. Its capacity to generate unlimited data offers promising pretraining opportunities for deep learning models.

\subsection{Discussion}  
The task of generating paired image-mask offers a significant opportunity in Surgical Data Science, particularly as a supplementary or alternative data source when access to real-world data is limited by ethical or regulatory constraints. While our study shows that models trained on SimGen's generated data alone do not outperform those trained on real data, the synthetic data still holds value in scenarios where real data is inaccessible or when trained on larger data. Specifically, SimGen can serve as a viable alternative for training AI models when real surgical data cannot be released, achieving moderate performance and offering a potential avenue for domain adaptation studies due to the distribution shift observed between real and synthetic data.

Beyond data scarcity issues, SimGen's ability to generate large amounts of synthetic data makes it a promising tool for pretraining deep learning models, establishing foundational patterns before fine-tuning on real data. This capability could extend to surgical simulation and educational applications, enhancing training resources for surgical trainees and professionals by providing realistic visual content. Additionally, the potential of SimGen to support domain shift studies opens new pathways for improving model generalization across different surgical datasets and environments.

Nevertheless, limitations persist. The model's performance can degrade on smaller or less diverse datasets, as seen with the Endoscape \cite{murali2023endoscapes}. The absence of controlled or conditioned image-mask generation reduces the flexibility of the system for specific clinical needs. High GPU demands for training may also may hinder accessibility for some institutions, particularly those with limited computational resources.
Synthetic data may introduce bias, moving forward, clinical experts involvement in the validation loop is crucial to ensure the generated data aligns with real-world surgical applications, 
addressing both ethical considerations and practical effectiveness.

\begin{figure}[!htbp]
    \centering
    \includegraphics[width=0.995\linewidth]{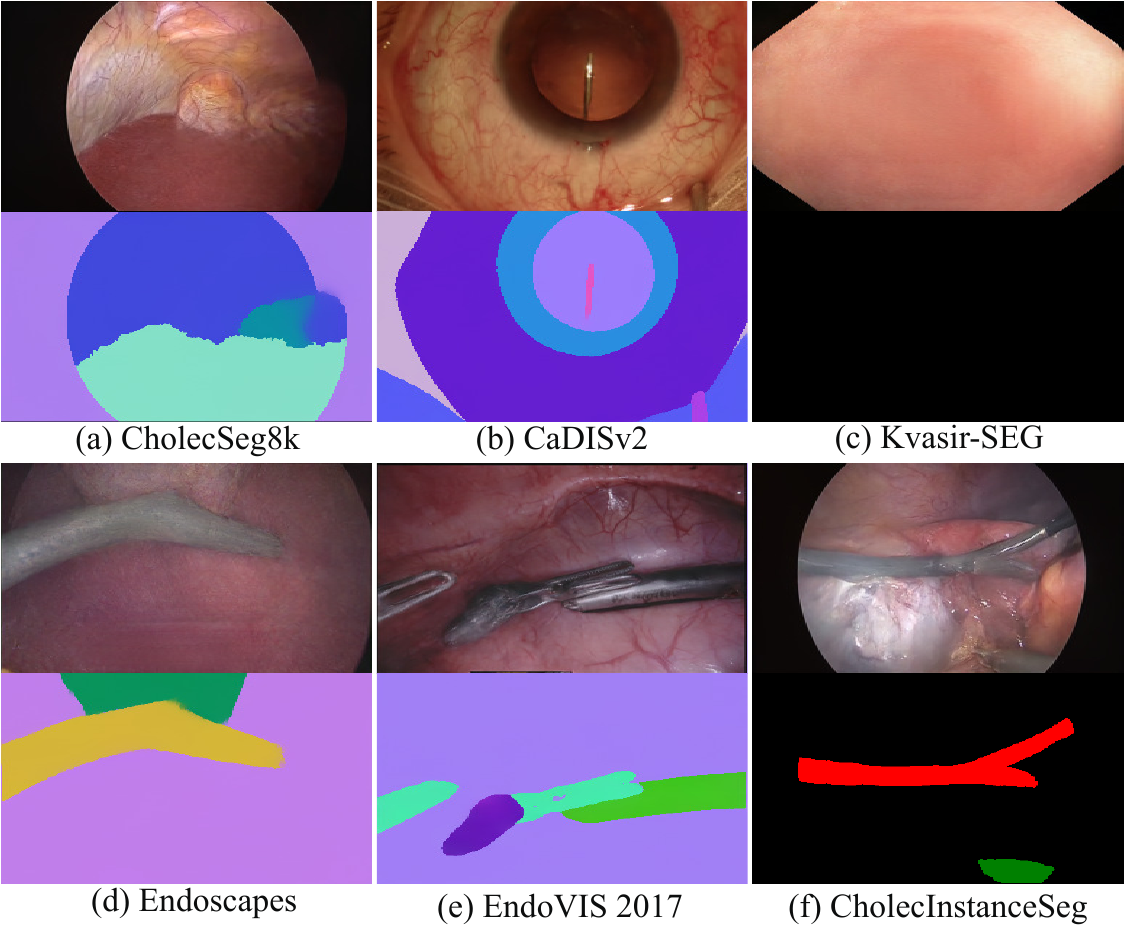}
    \caption{Examples of failure cases in generative model output for paired image-mask generation.}
    \label{fig:failure-cases}
\end{figure}

\subsection{Failure Cases}
We observe some failure case with SimGen. For analysis, we isolate a few failure cases across the datasets, as presented in Fig.~\ref{fig:failure-cases}. These occurrences are rare but provide valuable insights for further improvements. Below, we analyze specific instances across datasets:
\begin{enumerate}
    \item \textit{CholecSeg8k}: A few blurred images occasionally lead to slight inaccuracies in anatomical boundaries, affecting segmentation precision.
    \item \textit{CaDISv2}: Rare instances of incorrect tool localization, such as a "floating" needle in the pupil, result in minor deviations in the generated mask.
    \item \textit{Kvasir-SEG}: Occasionally, some anomalies are missed, and subtle over-smoothing is observed in the generated masks, reducing fine-grained details like vein structures in polyps.
    \item \textit{Endoscapes}: The limited size of the dataset (343 training pairs) sometimes leads to challenges in capturing intricate details of tools and anatomical structures, though this is infrequent.
    \item \textit{EndoVIS 2017}: In rare cases, the generated mask includes unexpected features, such as a "floating" grasper, slightly affecting tool placement.
    \item \textit{CholecInstanceSeg}: Few instances of tool merging in the generated mask (e.g., red regions) cause minor loss of tool instance identity.
\end{enumerate}

These occasional issues are not reflective of overall model performance but highlight opportunities for refinement. Increasing training data diversity and targeted improvements in the model could further mitigate these challenges and enhance robustness.


\section{Conclusion}
In this work, we address the challenge of generating high-quality paired image-segmentation masks, a critical bottleneck in Surgical Data Science. We introduce \textit{SimGen}, a diffusion-based model designed to simultaneously generate surgical images and their corresponding segmentation masks. This innovative approach offers data source without ethical concerns for surgical education, simulation, and other complex clinical applications.
Trained across five surgical datasets, SimGen  effectively produces realistic image-segmentation pairs, demonstrating improved FID, KID, sFID, and sKID scores compared to existing baselines. While models trained on SimGen-generated data do not outperform those trained solely on real data, we highlight the benefit in the complementary use of generated data when SimGen is trained on larger samples or broader class set of public real data, highlighting the potential for using synthetic data to support research and development, particularly in scenarios where real-world data is inaccessible or restricted.
We also demonstrate that \textit{SimGen} outperforms segmentation models in aligning masks with generated images for surgical simulation.
Looking ahead, \textit{SimGen} paves the way for creating synthetic datasets beyond segmentation masks, including image-bounding box pairs and video-segmentation data. Such advancements could significantly enhance surgical AI by providing scalable, high-quality synthetic data to improve model training and performance.


\subsection*{Acknowledgements:} This work was supported by French state funds managed within the Plan Investissements d’Avenir by the ANR under grants: ANR-20-CHIA-0029-01 (National AI Chair AI4ORSafety), ANR-22-FAI1-0001 (project DAIOR), ANR-10-IAHU-02 (IHU Strasbourg). 
This work was granted access to the servers/HPC resources managed by CAMMA, IHU Strasbourg, Unistra Mesocentre, and GENCI-IDRIS [Grant 2021-AD011011638R3, 2021-AD011011638R4].

\bibliographystyle{IEEEtran}
\bibliography{main-arxiv}

\end{document}